\documentclass[a4paper,conference]{IEEEtran}
\ifCLASSINFOpdf
\else
\fi

\usepackage{graphicx}
\usepackage{times}
\usepackage{latexsym}
\usepackage{times}
\usepackage{url}
\usepackage{latexsym}
\usepackage{amsmath}
\usepackage{amssymb}
\usepackage{latexsym}
\usepackage{graphicx}
\usepackage{multirow}

\usepackage{graphicx}
\usepackage{amsmath,graphicx}
\usepackage{times}
\usepackage{latexsym}
\usepackage{times}
\usepackage{latexsym}
\usepackage{caption}
\usepackage{subcaption}
\usepackage{comment}
\usepackage{times}
\usepackage{soul}
\usepackage{booktabs}
\usepackage{url}
\usepackage[utf8]{inputenc}
\usepackage{graphicx}
\usepackage{amsmath}
\usepackage{amsmath,graphicx}
\usepackage{times}
\usepackage{latexsym}
\usepackage{times}
\usepackage{latexsym}
\usepackage{comment}
\usepackage{times}
\usepackage{soul}
\usepackage{booktabs}
\usepackage{url}
\usepackage[utf8]{inputenc}
\usepackage{graphicx}
\usepackage{amsmath}
\usepackage{amssymb}
\usepackage{array}
\usepackage{booktabs}
\usepackage{multirow}
\usepackage{multicol}
\usepackage{caption}
\usepackage{xspace}
\usepackage{amssymb}
\usepackage{booktabs}

\usepackage{multirow}
\usepackage{multicol}
\usepackage{caption}
\usepackage{xspace}
\usepackage{amsmath}

\usepackage{tabularx}

\usepackage{latexsym}
\usepackage{multirow}
\usepackage{algorithm} 
\usepackage{algpseudocode} 
\usepackage{multicol}
\usepackage{comment}
\usepackage{hyperref}
\usepackage{amsmath}

\usepackage{tabularx}

\usepackage{graphicx}
\usepackage{subcaption}
\usepackage{url}
\DeclareSymbolFont{rsfs}{U}{rsfs}{m}{n}
\DeclareSymbolFontAlphabet{\mathscrsfs}{rsfs}

\begin{document}
%
\title{A Syntax Aware BERT for Identifying Well-Formed Queries in a Curriculum Framework}

\author{\IEEEauthorblockN{Avinash Madasu}
\IEEEauthorblockA{Department of Computer Science \\ University of North Carolina at Chapel Hill \\
Email: avinashm@cs.unc.edu}
\and
\IEEEauthorblockN{Anvesh Rao Vijjini} 
\IEEEauthorblockA{Department of Computer Science \\ University of North Carolina at Chapel Hill\\
Email: anvesh@cs.unc.edu}}

%


\maketitle

\begin{abstract}
A well formed query is defined as a query which is formulated in the manner of an inquiry, and with correct interrogatives, spelling and grammar. While identifying well formed queries is an important task, few works have attempted to address it. In this paper we propose transformer based language model - Bidirectional Encoder Representations from Transformers (BERT) to this task. We further imbibe BERT with parts-of-speech information inspired from earlier works. Furthermore, we also train the model in multiple curriculum settings for improvement in performance. Curriculum Learning over the task is experimented with Baby Steps and One Pass  techniques. Proposed architecture performs exceedingly well on the task. The best approach achieves accuracy of 83.93\%, outperforming previous state-of-the-art at 75.0\% and reaching close to the approximate human upper bound of 88.4\%.
\end{abstract}


%
\IEEEpeerreviewmaketitle

\section{Introduction}
Search Engines like Google and Bing, and E-commerce websites like Amazon and Flipkart rely on well-formed user queries to retrieve results expected by the user. However, query usage can vary from user to user though having the same intent. This can be understood as every user has a particular usage of words, grammatical knowledge and its structure \cite{barr-etal-2008-linguistic}. Though a query can be formed in different ways, its intent remains the same making it difficult for the search engines to understand. The query language can deviate from Natural Language (NL) grammatical rules such as queries made of only nouns, pronouns and verbs, spelling mistakes. This makes standard Natural Language Processing (NLP) tools trained on well-formed data difficult to understand these queries. Furthermore, inherent sensitiveness of the language can create ambivalent semantic outputs for the same query debasing user experience.

Identifying if a query is well formed helps in restructuring queries better understood by the search engines. Early attempts were made to classify a query into informational and not-informational using supervised and unsupervised methods \cite{baeza2006intention}. Segmenting a query into sub-parts using nouns improved precision of document retrieval \cite{bergsma2007learning}. As the time progressed, queries became much more sophisticated hence requiring advanced NLP techniques and datasets. In this direction forward, a new annotated dataset is introduced with natural language questions and the probabilities of being well-formed \cite{faruqui2018identifying}. 

Transformer based language models achieved State-of-the-art (SotA) performance on Natural Language Understanding (NLU) tasks. However, a significant breakthrough was achieved by BERT \cite{devlin2018bert}, a Bi-directional Transformer Encoder pretrained on large amounts of data and fine-tuned for downstream NLU tasks. Since then, BERT is being used to perform fine-tuning on other tasks such as Domain Adaptation \cite{ma-etal-2019-domain}, Question Answering \cite{osama-etal-2019-question}.

Curriculum learning introduced and formulated by \cite{bengio2009curriculum} is a method in which neural networks are trained using easy samples at first and in the succeeding phases, difficult samples are introduced to the model. Its inspiration lies in cognitive science theories \cite{skinner1958reinforcement,krueger2009flexible} which propose for humans to acquire a skill, they are first provided easier variants of the task following increased difficulty. Effectiveness of Curriculum Learning has been explored in Natural Language Understanding tasks such as Question Answering \cite{sachan2016easy} and Natural Answer Generation \cite{liu2018curriculum} and more recently in the broad range of GLUE tasks as well \cite{xu2020curriculum}. \cite{cirik2016visualizing} proposed Baby Steps and One Pass curriculum techniques using sentence length as a curriculum strategy for training LSTM \cite{hochreiter1997long} on Sentiment Analysis. A tree-structured curriculum ordering based on semantic similarity is proposed by \cite{han2017tree}. \cite{rao2020sentiwordnet} propose a auxiliary network which is first trained on the dataset and used to calculate difficulty scores for the curriculum ordering. Some of these works \cite{cirik2016visualizing,han2017tree,rao2020sentiwordnet} have suggested that Baby Steps performs better than One Pass. We perform experiments using both the techniques. While the idea of curriculum remains same across these works, the strategy itself to decide sample ordering is often tough to decide. In this work, we exploit the query well-formedness probability scores given in the query well-formedness dataset \cite{faruqui2018identifying} to design our curriculum ordering. In doing so we forego the experiments required to choose a curriculum ordering and furthermore, build a strategy of difficulty which agrees with the perspective of annotators and domain experts.

The overall contributions of our paper are as follows:
\begin{itemize}
    \item We establish a new state-of-the-art performance in query well-formedness identification utilising BERT.
    \item Proposed framework utilizes query well-formedness probability scores to determine a curriculum ordering for training. Samples are ordered in this strategy based on how difficult is it to differentiate them between well-formed or poorly-formed.
    \item Proposed architecture also infuses syntactic formation in the form of Parts-of-Speech in a manner that helps the model achieves better performance.
\end{itemize}

In Section \ref{sec:related_work}, we explain previous SotA approaches which tackle Query well-formedness identification and also, curriculum learning for text classification works. Section \ref{sec:data} explains the query well-formedness dataset. Section \ref{sec:approach} explains our Approach. Section \ref{sec:arch} provides our architecture in a detailed manner including the curriculum algorithm. Section \ref{sec:baselines} lists down the baselines for comparison and other recent architectures proposed for this task. Section \ref{sec:results} explains our results and Section \ref{sec:con} concludes our work.

\section{Related Work}
\label{sec:related_work}
While the importance of query well formedness prediction is immense, few works have addressed it. Faruqui \& Das \cite{faruqui2018identifying} propose solutions for the task using character, word and parts-of-speech features. They propose  a feed forward neural network with two hidden layers which takes word, character and POS n-gram from  SyntaxNet POS tagger\cite{alberti2015improved}, all concatenated into the input layer. Their approach shows a significant improvement by including POS features.

Faruqui \& Das \cite{faruqui2018identifying} further identify that a query is often not well formed because of its ungrammatical nature. POS sequences often encode syntactic information and hence help in identifying ungrammatical structures in poorly formed queries.

Syed \cite{syed2019inductive} proposed the popular ULMFiT\cite{howard2018universal} architecture for Query well formedness. They employ a framework of Inductive Transfer Learning to train their architecture. In this framework, the ULMFiT architecture which consists of stacked AWD-LSTMs\cite{merity2017regularizing} is first pretrained in a language model setting on open domain data. It is followed by further language model fine tuning on the query well-formedness dataset and finally the actual classification down stream task. Language model pre-training helps the model learn and understand the nature, semantics and structure of regular language. As a consequence, when the model is tasked with differentiating between grammatical and ungrammatical construct, it is able to utilize the language model information.

Furthermore, as \cite{faruqui2018identifying} suggest, well formedness of a query relies heavily on grammatical structure and how semantically sound the query is. This parallels the dataset of Corpus of Linguistic Acceptability CoLA \cite{warstadt2019neural}. CoLA is a dataset intended to gauge at the linguistic competence of models by making them judge the grammatical acceptability of a sentence. CoLA is also part of the popular GLUE benchmark datasets for Natural Language Understanding \cite{wang2018glue}. The transformer \cite{vaswani2017attention} encoder architecture BERT \cite{devlin2018bert} is known for outperforming previous GLUE SotAs, including CoLA. Hence we utilise BERT as our base architecture.

\begin{table*}[h]
\begin{tabularx}{\textwidth}{l X l}
    \hline
    Label &  Example & Query well-formedness probability\\
    \hline
     & 1. 1961 penney worth ? & 0.0 \\
    
    Easy  & 2. How many years of college do you have to do to be an airline pilot ? & 1.0\\
    
     & 3. 8OZ IN 1000ML ? & 0.0 \\
    \hline
      & 1. Why use nautical mile unit ? & 1.0  \\

    Medium   & 2.Where is Ann Bates graive ?& 0.2  \\

       & 3. What is turkey population ? &0.2 \\
     \hline
    & 1. How many potatoes is 550 grams ? &0.8 \\

    Hard     & 2.Can hamerhead sharks eat ? &0.4 \\

       & 3. What is the worlds easiest drawing ? &0.6 \\
     \hline
     \hline
\end{tabularx}
  \caption{Examples of Difficult and Easy samples according to the proposed curriculum.}
  \label{tab:data_examples}
\end{table*}
\section{Dataset}
\label{sec:data}
To evaluate our proposed model, we perform experiments on query well-formedness dataset \cite{faruqui2018identifying}. This dataset has collection of user generated queries and their associated annotator tagged query well-formedness scores. For the dataset building process, annotators consider a query well-formed if the words are spelled properly, the  sentence syntax is syntactically and grammatically sound and if the sentence is interrogative in the pragmatic and discourse sense. We use the same train, dev and test splits and 0.8 is the threshold used for classification. Unless specified, the results reported from our experiments are averaged across 5 runs. Example samples from this dataset along with their associated well-formedness scores can be observed in Table \ref{tab:data_examples}.
\begin{table}
\centering
\begin{tabular}{l|l|l|l} 
\hline
 Difficulty Set& Positive & Negative & Total  \\ 
\hline
\hline
Hard &2578 &2002 &4580 \\ 
\hline
Medium &4189 &4860 & 9049\\ 
\hline
Easy & 4189&3773 &7962 \\ 
\hline
\end{tabular}
\caption{Dataset Details.}
\label{tab:data_details}
\end{table}


\section{Proposed Approach}
\label{sec:approach}
\begin{figure*}
\centering
  \includegraphics[width=12cm]{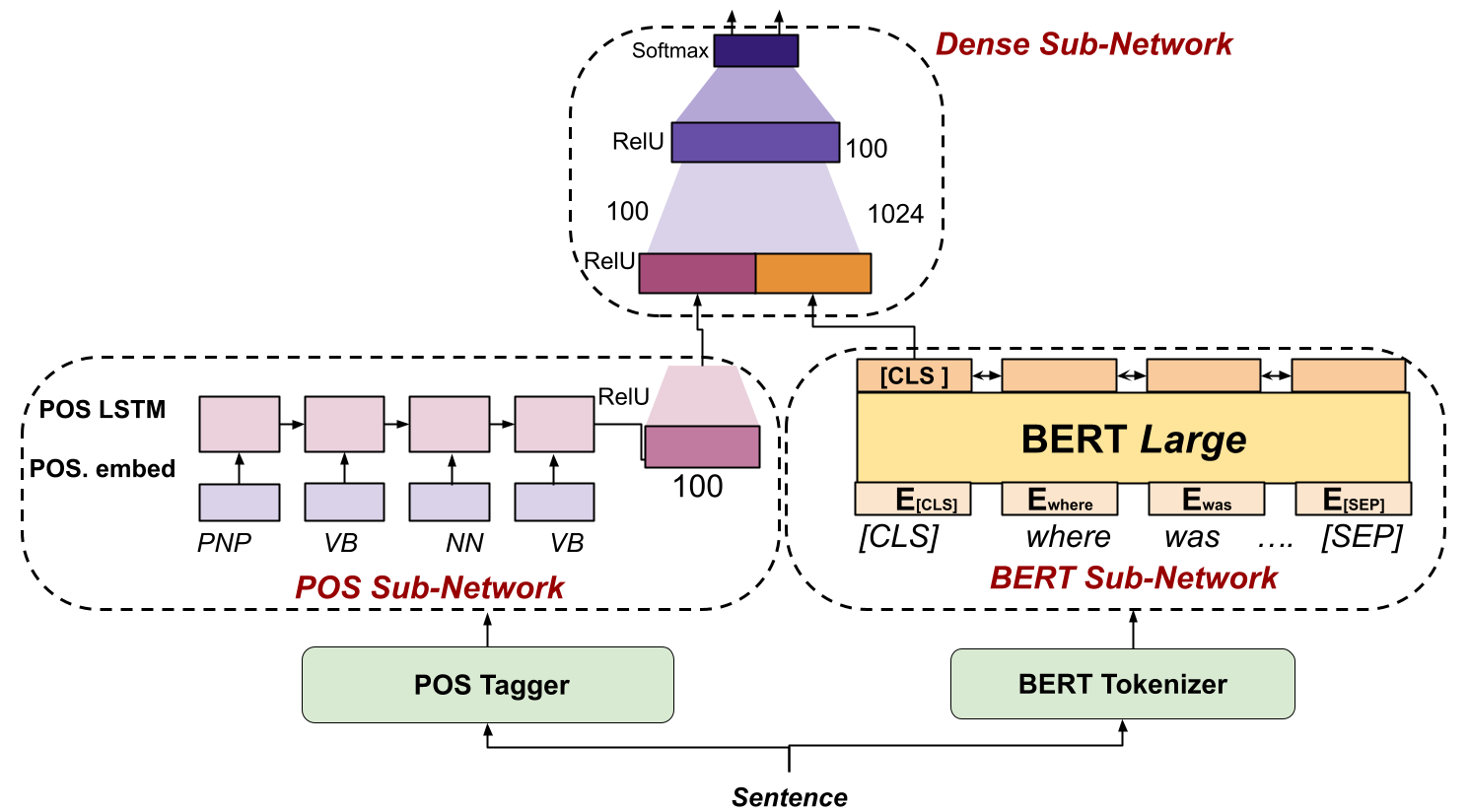}
  \caption{Proposed POS-BERT Architecture}
  \label{fig:architecture}
\end{figure*}

In this section, we describe architecture of the proposed model. The architecture of the proposed model is shown in figure \ref{fig:architecture}.
\subsection{BERT}
Bidirectional Encoder Representations from Transformers(BERT) \cite{devlin2018bert} is a masked language model trained on a large corpora. A sentence is added with a special token (CLS) at the beginning and is passed into pretrained BERT model. It tokenizes the sentence with a maximum length of 512 and outputs a contextual representation for each of the tokenized words. There are variants of pretrained BERT depending upon hyper-parameters of the model.
BERT-Base Uncased consists of 12 transformer encoders and output from each token is a 768 dimension embedding. BERT-Large Uncased is a stack of 24 transformer encoders and the output from each token is of 1024 dimension.
\subsection{POS-LSTM}
Although BERT is capable of learning underlying grammatical structure \cite{tenney2019bert}, modelling their structures can significantly improve performance. Infusing BERT with syntactic information in terms of Parts-of-Speech tags is done in previous works as well, such as \cite{sundararaman2019syntax}. However, they align POS information with BERT time steps. This leads to certain POS tags to be segmented since BERT relies on WordPiece tokenization \cite{kudo2018sentencepiece}. WordPiece segments out-of-vocabulary words into subwords that are observed in vocabulary. Aligning POS tags to WordPiece segments will affect the POS n-gram information negatively. Hence, we employ a separate LSTM trained on only POS tags\footnote{We use the NLTK toolkit for POS tagging: https://www.nltk.org/book/ch05.html}, thereby keeping their sequence and structure independent of BERT's tokenization. Significant improvement in LSTM performance by modelling them on syntactic information has been observed by previous works such as \cite{kuncoro2018lstms}.

\begin{figure*}
\centering
        \begin{subfigure}[b]{0.5\textwidth}
                \centering
                \includegraphics[width=5cm]{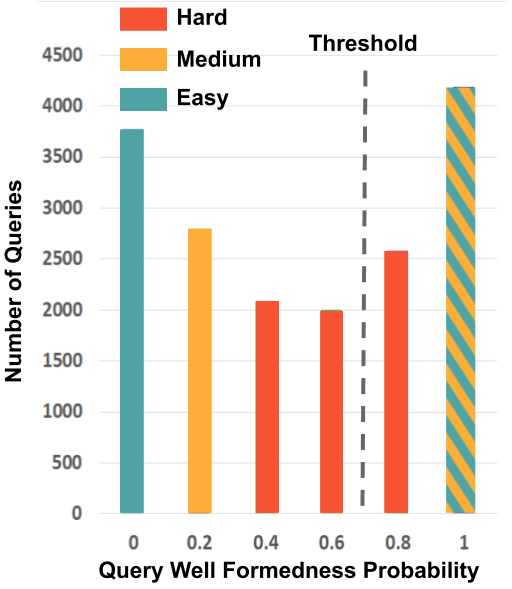}
                \caption{The distribution of the annotated question according to well-formedness probability as per \protect\cite{faruqui2018identifying} and our curriculum sets described.}
                \label{fig:dist}
        \end{subfigure}%
        \begin{subfigure}[b]{0.5\textwidth}
                \centering
                \includegraphics[width=5cm]{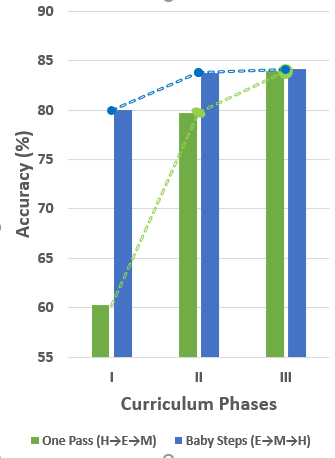}
                \caption{One Pass in Curriculum Ordered training}
                \label{fig:curr}
        \end{subfigure}
        \caption{}
\end{figure*}

\subsection{Effective utilization of query well-formedness probability for curriculum learning}
The significance of the probability score for a query was often overlooked by the previous literature. They considered 0.8 as the threshold and performed binary classification of the query being well-formed or not. It is intrinsically assumed that queries belonging to a particular class have the same characteristics. However, this assumption is often not true. For example, assume queries $q_{1}$, $q_{2}$, $q_{3}$ and $q_{4}$ with probability scores 0.0, 0.6, 0.8 and 1.0 respectively. $q_{1}$ and $q_{2}$ belong to same class whereas $q_{3}$ and $q_{4}$ belong to different class. $q_{1}$ is easy to distinguish from $q_{4}$ because they are distant from threshold and hence are easy to train. However, $q_{2}$ and $q_{3}$ are hard to train because they are close to threshold and are quite difficult to distinguish. Therefore, we categorize training samples into hard, medium and easy based on their scores. The category division is shown in Figure \ref{fig:dist}. We employ curriculum learning to train these categories so that adequate importance is given to hard samples.

\section{Architecture Details}
\label{sec:arch}
Let $X$ be the sentence where $X = \{x_{1},x_{2},...,x_{m}\}$, where $m$ is the sentence length and varies for each sentence.
\subsection{BERT sub-network}
Sentence $X$ is passed through BERT pretrained model which gives output from the last layer.
\begin{equation}
    H^{L}_{M} = BERT(X)
\end{equation}
where $M$ is the maximum token length of the BERT and $L$ is the number of encoder layers in BERT. The context embedding from [CLS] token in the last layer is $H^{L}_{0}$. 
\subsection{POS sub-network}
Let $P$ be the Parts-of-Speech tags for the sentence $X$ where $P = \{p_{1},p_{2}.....,p_{m}\}$ and $m$ is the sentence length. The embeddings for POS tags are initialized randomly and are updated during training. Let $E$ be the embedding matrix for POS tags $P$.
\begin{equation}
    E = emb(P)
\end{equation}
This embedding vector is trained using LSTM through which grammatical structure is modelled. Let $h_{t}$ be the output from the final time-step of LSTM.
\begin{equation}
    h_{t} = LSTM(E)
\end{equation}
The output from LSTM is passed through a fully connected layer.
\begin{equation}
    C = f(h_t \cdot W_{c})
\end{equation}
where $f$ is the activation function.

\subsection{Dense sub-network}
The outputs $H^{L}_{0}$ and $C$ from BERT sub-network and POS sub-network respectively are concatenated.
\begin{equation}
    T =  [H^{L}_{0}, C]
\end{equation}
The concatenated output $T$ is passed onto a fully connected layer.
\begin{equation}
    S = f(T \cdot W_{t})
\end{equation}
The output $S$ from the fully connected layer is passed onto fully connected softmax layer.
\begin{equation}
    O = g(S \cdot W_{s})
\end{equation}
where $g$ is the softmax activation function.

\subsection{Hyper-parameters}
We consider 44 as the maximum sentence length for BERT tokenization and 20 as the maximum sentence length for POS tags. 768 is the context embedding dimension for BERT Base Uncased whereas 1024 is the size of context embedding for BERT Large Uncased. ReLU is the activation function used in all the fully connected layers except in final layer. A dropout \cite{JMLR:v15:srivastava14a} of 0.5 is applied on the concatenated output. Batchsize used is 50 and the model is trained for 10 epochs. AdamW \cite{loshchilov2017decoupled} is used as the optimizer with a learning rate of 2e-5 and epsilon value 1e-8. We selected the model that performed best on validation data. Early Stopping with a patience of 5 is used if the model accuracy is not improving on validation data.
\subsection{Curriculum Training}
\begin{algorithm}
	\caption{One Pass Curriculum} 
	\begin{algorithmic}[1]
	\Procedure{OP-Curriculum}{$f_{w},D$}
		\State Obtain Ordering $(D_{1},D_{2},D_{3})$ 
			\For{$D_{i} = (D_{1},D_{2},D_{3})$}
				\State Train $f_{w}$ on $D_{i}$ 
			\EndFor
			\State Test $f_{w}$
    \EndProcedure
	\end{algorithmic} 
\label{algo:one}
\end{algorithm}
\begin{algorithm}
	\caption{Baby Steps Curriculum} 
	\begin{algorithmic}[1]
	\Procedure{BS-Curriculum}{$f_{w},D$}
		\State Obtain Ordering $(D_{1},D_{2},D_{3})$
		\State $\mathscrsfs{D}$ = $D_{1}$
		\State Train $f_{w}$ on  $\mathscrsfs{D}$
			\For{$D_{i} = (D_{2},D_{3})$}
			    \State $\mathscrsfs{D}$ = $\mathscrsfs{D}$ $+$  $D_{i}$ 
			    \State Train $f_{w}$ on  $\mathscrsfs{D}$
			\EndFor
			\State Test $f_{w}$
    \EndProcedure
	\end{algorithmic} 
\label{algo:baby}
\end{algorithm}

We propose to use two curriculum methods for training our model:  One Pass and Baby steps. They are described in Algorithms \ref{algo:one} and \ref{algo:baby} respectively. Here $f_{w}$ is the classification model, and $D$ is the whole Dataset made up of sets based on difficulty $D_{1}$,$D_{2}$ and $D_{3}$. In Baby Steps, we first train the model using easy samples. In the second phase, model is trained with both easy and medium samples. In the final phase, model is trained with the entire dataset. In One Pass, model is trained using hard, easy and medium samples subsequently. Note that in one pass samples which used to train in the previous phase are not added to the samples used in the current phase.
\section{Baselines}
\label{sec:baselines}
We compare the proposed model to previous State-of-the-Art (SotA) architectures.
\subsection{Question Word Classifier}
If the query in the test set begins with a question classify it as well-formed \cite{syed2019inductive}.
\subsection{Majority Class Prediction}
In this model, we classify all the queries into the majority class of the test set \cite{syed2019inductive}.
\subsection{Word BiLSTM}
A Bidirectional LSTM is trained with one-hot vectors as the input from the words of a query. The output from final time-step is passed onto a softmax classifier to perform binary classification \cite{syed2019inductive}.
\subsection{word-1,2 char-3,4 grams}
A feed-forward neural network is trained with word-1,2 grams and character-3,4 grams as input features. All the features are concatenated to form the input to the network. Every feature is represented as a real-valued embedding \cite{faruqui2018identifying}. 
\subsection{word-1,2 POS-1,2,3 grams}
In this model, in addition to word-1,2 grams, POS features are extracted using SyntaxNet POS tagger \cite{alberti2015improved}. All these features are concatenated to form input to the feed-forward neural network.
\subsection{word-1,2 char-3,4 POS-1,2,3 grams}
This model's input is a concatenation of features such as word-1,2 grams, character-3,4 grams and POS-1,2,3 grams trained using feed-forward neural network.
\subsection{Inductive Transfer Learning (ITL)}
In this model, Pretrained Language Model ULMFiT \cite{howard2018universal} is fine-tuned onto the well-formedness query dataset. The Fine-tuned Language Model is used to train a classifier by adding a softmax layer on the top of ULMFiT Language Model \cite{syed2019inductive}.

\begin{table*}
\centering
\begin{tabular}{l|l|l} 
\hline
 Model & Acc & F1  \\ 
\hline
\hline
Question Word Classifer \cite{syed2019inductive}&54.9 &- \\ 
\hline
Majority Class Prediction \cite{syed2019inductive} & 61.5& - \\ 
\hline
Word BiLSTM Classifier \cite{syed2019inductive} &65.8&- \\ 
\hline
word-1,2 char -3,4 grams \cite{faruqui2018identifying} &66.9& - \\ 
\hline
word-1,2 POS -1,2,3 grams \cite{faruqui2018identifying} & 70.7 & - \\ 
\hline
word-1,2 char-3,4 POS-1,2,3 grams \cite{faruqui2018identifying}&70.2 &  - \\ 
\hline
ITL \cite{syed2019inductive}& 75.0& -  \\
\hline
\hline
BERT-Base & 78.56& 77.4 \\
\hline
BERT-Large & 82.55&   82  \\ 
\hline
BERT-Large + POS & 83.04 & 82.2    \\
\hline
BERT-Large + POS + Baby Steps Curriculum&83.39 &   \textbf{83}  \\
\hline
BERT-Large + POS + One Pass Curriculum&\textbf{83.93} &  \textbf{83}   \\
\hline
Approx. human upper bound \cite{faruqui2018identifying} & 88.4 & - \\
\hline
\end{tabular}
\caption{Comparison of proposed approach with recent architectures. FT denotes Fine-Tuning}
\label{tab:results}
\end{table*}

\section{Results and Discussion}
\label{sec:results}
Table \ref{tab:results} captures the results of proposed approach with SotA approaches along with ablation studies for the proposed approach in terms of Accuracy and F1 score. As we see in Table \ref{tab:data_details}, the data is imbalanced, in such cases  F1 is a much more precise metric for comparison. However previous approaches have only measured Accuracy for the task, and we utilize the same for comparison with these works. 

\begin{figure*}[!htbp]
\centering
  \includegraphics[width=\linewidth]{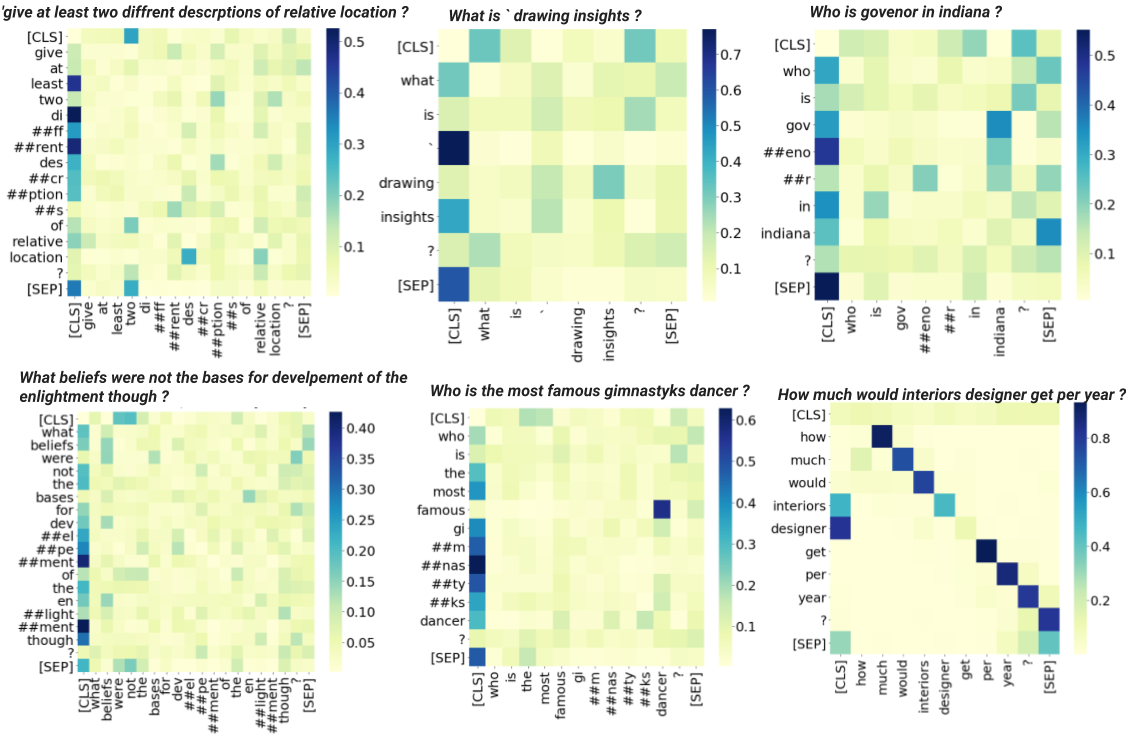}
  \caption{BERT's Attention visualized. All visualizations but the last one are from the $10^{th}$ head of the first layer. The last attention is from the $12^{th}$ head of the same layer.}
  \label{fig:viz}
\end{figure*}

\subsection{BERT-Base outperforms previous works} 

Even without any additional inputs, we observe that BERT-Base Uncased and BERT-Large Uncased outperforms previous approaches. Both the BERT methods and ITL are language models pretrained on huge amounts of open domain data which infuse them with general language structure. However, the self-attention characteristic of BERT helps it to attend to positions which cause the sentence to be poorly formed such as incorrect spellings or ungrammatical structures thereby improving performance with such identifications. To further investigate the importance of self-attention the task of query well-formedness identification we visualize BERT's attention in Figure \ref{fig:viz}. Firstly, we notice that BERT's attention heads attend to different ideas relevant for the classification task. These ideas especially include incorrect spellings and ungrammatical structures. Works such as \cite{clark2019does} establish that BERT's heads attend on linguistic notions of syntax. We find that BERT is able to identify and attend to incorrect spellings exceptionally well, even for a case where there are multiple incorrect spellings in a sentence (``develpement" and ``enlightment"). Previous approaches \cite{faruqui2018identifying,syed2019inductive} would treat an incorrect spelling as simply an out-of-vocabulary word. WordPiece tokenizer allows it to break incorrect spellings and attend to improbable sub-word sequences helping in predicting poorly formed queries. We found certain heads focusing on syntax and grammar as well. An example is visualized where BERT is attending on a noun phrase with incorrect number information (``interiors designer")

\subsection{Importance of POS level knowledge}

We see that the introduction of POS LSTM to BERT, further improves the result. This reinforces the idea that parts-of-speech information is critical in capturing unsound grammar \cite{faruqui2018identifying}. However, the improvement over the introduction of POS to BERT is less than the introduction of POS-level information to \cite{faruqui2018identifying}'s architecture. This shows that while POS helps in the task, BERT alone also has good knowledge of syntax and grammar, an idea widely popular in previous works as well \cite{tenney2019bert}.

\subsection{Curriculum Strategies}
Finally, among One Pass and Baby Steps curriculum strategies, we see that both perform exceedingly well, with results very close to approximate human upper bound as defined by Faruqui \& Das \cite{faruqui2018identifying}. One Pass is more efficient since it only observes a sample once, unlike Baby Steps which repeatedly sees samples from previous phases along with current phase's samples. In our experiments we observe that both the approaches perform the best under different orderings of samples. Baby Steps performed best in an Easy to Hard setting like in a typical curriculum. However, one pass performed best with Hard first, followed by Easy to Medium setting. Previous works \cite{cirik2016visualizing,han2017tree,rao2020sentiwordnet} have shown that just an easy first approach does not work for One Pass Strategies. In the proposed curriculum setting difficulty of samples is defined based on the well formedness probability scores. The intuition behind categorizing samples as difficult is whether an annotator would get confused in distinguishing the sample between well formed or poorly formed. This can be observed in Table \ref{tab:data_examples} which shows examples of Easy and Hard samples according to the proposed difficulty strategy. It's fairly easy to distinguish between the well formed and poorly formed queries in the Easy samples and conversely for the Hard subset. Interestingly, in Figure \ref{fig:curr} we see that such a definition of difficulty agrees with the proposed approach as well, since we see that in One Pass $1^{st}$ phase, training on hard samples leads to performance as bad as majority class prediction. This implies that the proposed BERT model is approaching the task of query well-formedness in a human like manner which explains the near human level performance. The accuracy scores for one pass in each run are as follows: 83.7, 83.85, 84.1, 83.97 and 84.03. The F1 scores for one pass in each run are 82.8, 82.97, 83.07, 83.11 and 83.05.

\section{Conclusion}
\label{sec:con}
We propose BERT for the task of identifying well formed queries. Visualizations of BERT's attentions further shed light behind its effectiveness. We further enrich our performance by utilizing Parts-of-Speech information in proposed architecture. POS tags are passed through a LSTM to learn POS sequences which capture grammatical information relevant for the task. Finally, we use the query well formedness probability scores to derive a curriculum order for training the proposed approach. The curriculum learning is done in one-pass and baby steps setting in orders that suit them best. Proposed approach achieves near human performance and outperforms previous approaches.



\bibliographystyle{IEEEtran}
\bibliography{IEEEabrv.bib}

\end{document}